\pdfoutput=1
\documentclass[runningheads]{llncs}

\usepackage{amssymb}
\usepackage{amsmath}
\usepackage{url}
\usepackage{graphicx}
\usepackage{multirow}
\usepackage{wrapfig}
\usepackage{hyperref}
\usepackage{color}
\usepackage{bbm}
\usepackage{graphicx}
\usepackage[font=small,labelfont=bf]{caption}

\begin{document}

\title{PRSNet: Part Relation and Selection Network for Bone Age Assessment }

\author{Yuanfeng Ji\inst{1,2}\and 
	Hao Chen\inst{1} \and
	Dan Lin\inst{3} \and
	Xiaohua Wu\inst{1} \and
	Di Lin\inst{2}\thanks{Di Lin is the corresponding author of this paper.}}
%1{Yuanfeng,Ji}
%2{Hao,Chen}
%3{Dan,Lin}
%4{Xiaohua,Wu}
%5{Di,Lin}

%
\authorrunning{F. Author et al.}
\institute{Imsight Medical Technology, Co., Ltd., China \and
	Shenzhen University, China \and
	Dept. of Computer Science and Engineering, The Chinese University of Hong Kong, Hong Kong SAR, China\\}

\maketitle
\begin{abstract}
Bone age is one of the most important indicators for assessing bone's maturity, which can help to interpret human's growth development level and potential progress. In the clinical practice, bone age assessment (BAA) of X-ray images requires the joint consideration of the appearance and location information of hand bones. These kinds of information can be effectively captured by the relation of different anatomical parts of hand bone. Recently developed methods differ mostly in how they model the part relation and choose useful parts for BAA. However, these methods neglect the mining of relationship among different parts, which can help to improve the assessment accuracy. In this paper, we propose a novel part relation module, which accurately discovers the underlying concurrency of parts by using multi-scale context information of deep learning feature representation. Furthermore, based on the part relation, we explore a new part selection module, which comprehensively measures the importance of parts and select the top ranking parts for assisting BAA. We jointly train our part relation and selection modules in an end-to-end way, achieving state-of-the-art performance on the public RSNA 2017 Pediatric Bone Age benchmark dataset and outperforming other competitive methods by a significant margin.
\end{abstract} 
\section{Introduction}
\label{sec:intro}

Bone age assessment (BAA) requires to interpret the maturation of bone, playing an important role in understanding the growth of human. It is utilized in an array of scenarios, such as the diagnosis and treatment of disorder of body.
Generally, in clinical diagnosis, radiologists estimate the bone age by using the Graulich-Pyle(G-P) \cite{greulich1959radiographic} and Tanner-Whitehouse(T-W) methods \cite{tanner1975assessment}, which rely on expertise at the cost of tremendous time for observing each sample. However, due to the complex pattern of bones, different experts may provide various observations, easily leading to problematic judgements for the down-stream diagnosis and treatment. Thus, recent methods \cite{davis2012automated,thodberg2009bonexpert} incorporate more effective computer-aided system to assist BAA. For example, the BoneXpert\cite{thodberg2009bonexpert} diagnosis system has been approved and applied in various countries. Yet, these systems require too expensive process to generate high-quality images for judging bone ages.

The latest methods \cite{bae2018improved,halabi2018rsna,iglovikov2018paediatric,spampinato2017deep,wu2018residual} borrow the success of deep networks in medical image analysis \cite{bejnordi2017diagnostic}\cite{dou2016automatic}, and apply deep learning framework in BAA. To address the large variation of bones with respect to positions, sizes and shapes, Pan et al. \cite{halabi2018rsna} and Iglovikov et al. \cite{iglovikov2018paediatric} partition a bone structure into parts, and select a single part as input to train network individually. It leads to a more focused learning of invariant features, which are discriminative to bone ages. Wu et al. \cite{wu2018residual} uses the attention module to detect key parts of the hand bone. Bae et al. \cite{bae2018improved}, David.B et al. \cite{larson2017performance} and Spampinato et al. \cite{spampinato2017deep} further extract features on multiple parts to yield richer information. However, the previous methods neglect the relationship between parts of a bone structure, which is important to select useful parts for BAA.

In this paper, we advocate the idea of building relationship between parts of the hand bone and selecting discriminative parts for BAA. Given the hand bone in X-ray images, we propose a \emph{Part Relation Module} which connects strong-correlated parts. It enables the direct communication between parts, embedding more effective context information in part representations. Based on the part relationship, we employ a \emph{Part Selection Module} to harness useful parts for the final estimation of bone age. Note that the part selection is done by self-supervision, without the requirement of heavy labelling effort. More importantly, the relation and selection modules form an end-to-end framework, jointly distilling the features for BAA. This enables our approach to provide more details, i.e., part relation map and part selection results, which are critical to medical analysis. Our method achieves state-of-the-art result on the public benchmark, i.e., RNSA pediatric bone age dataset. It demonstrates the effectiveness of our approach.

\begin{figure*}[t!]
\centering
\includegraphics[width=\linewidth]{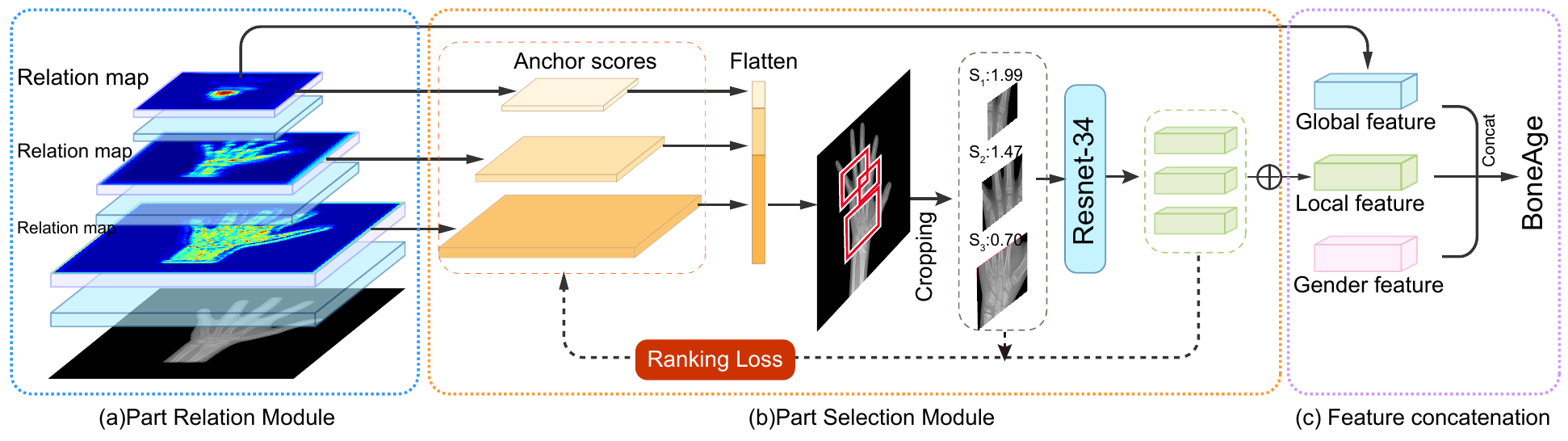}
\caption{Our PRSNet consists of the part relation module (a), selection module (b). Finally, we perform the feature concatenation (c) to include global feature, local feature and gender feature for estimating the bone age.}
\vspace{-0.2in}
\label{fig:overview}
\end{figure*}

\section{Part Relation Module}
Compared to individual parts of the hand bone, part relation provides richer information for understanding the importance of different parts, and yields more accurate information for BAA. But recent methods use independent parts to learn assessment models, which are oblivious to part relations. The latest works compute global representation on the whole image, and implicitly models part relation. However, the whole image is insensitive to positions, sizes and shapes of parts, making it difficult to construct effective representations.

In this paper, we propose the part relation module to discover the useful relationship between parts for BAA. As illustrated in Fig. \ref{fig:overview}(a), given an X-ray image, we use a CNN to compute different levels of convolutional feature maps (see the blue blocks). Each feature map is used to compute a corresponding relation map for modelling part relation. In the relation map, we produce high responses for strongly-correlated parts of the corresponding hand bone, while suppressing irrelevant regions. At each level, we use the convolutional feature map and the associated relation map to produce the context representation of correlated parts. We use all levels of context representations to provide more detailed information, for precisely scoring parts that have variant spatial and appearance properties.

\begin{figure*}[t!]
\centering
\includegraphics[width=\linewidth]{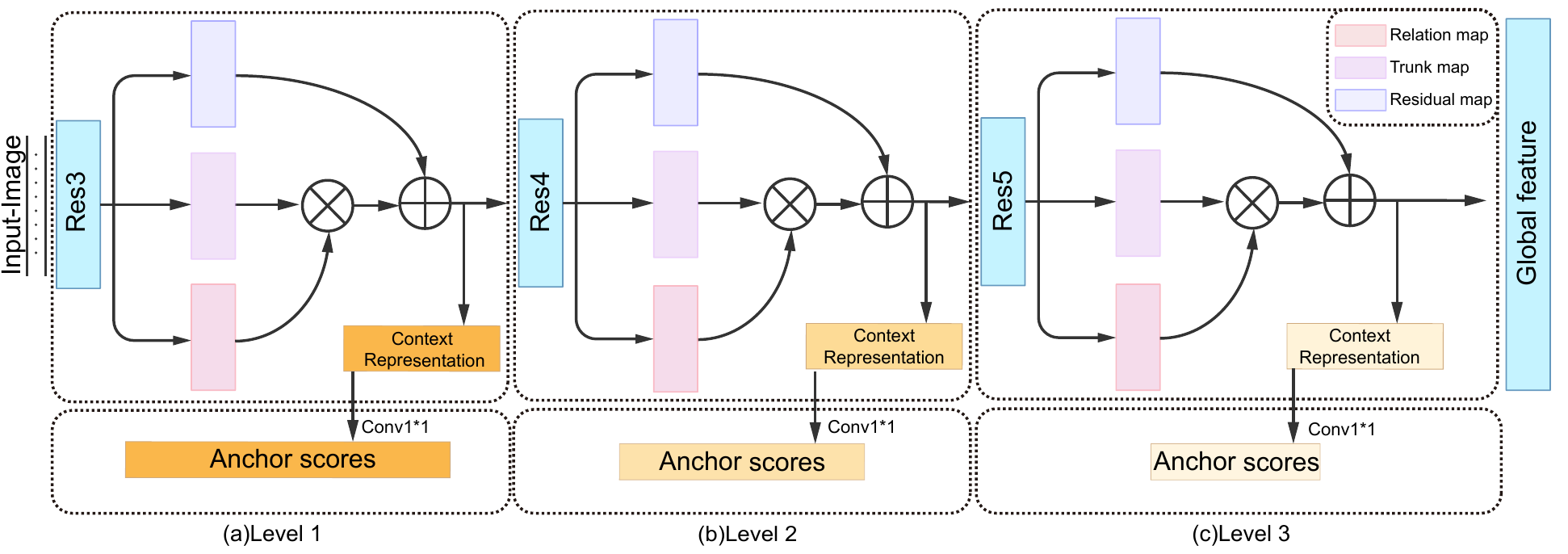}
\caption{The architecture of the part relation module. At each level, the convolutional feature map of backbone network undergoes different convolutional operations to yield relation map, trunk map and residual map. The new feature maps are used to compute the context representation, which is fed to estimate anchor scores in the part selection module (see Fig.~\ref{fig:anchor}).}
\vspace{-0.2in}
\label{fig:part_relation}
\end{figure*}

More formally, we use a backbone ResNet-34~\cite{he2016deep}to produce a set of convolutional feature maps $\{X_i\}$. The feature map $X_i$ is output by the residual block at the $i^{th}$ level. As shown in Fig.~\ref{fig:part_relation}, we apply different $1 \times 1$ convolutional operations on the feature map $X_i$ to produce the relation map $R_i$, trunk map $T_i$ and residual map $D_i$, respectively. Then we compute the context representation of correlated parts as:
\begin{equation}
F_i = T_i \cdot \sigma(R_i) + D_i,
\label{eq:context}
\end{equation}
where $\sigma$ is sigmoid activation function. In Eq.~\eqref{eq:context}, we apply the sigmoid activation function to the relation map $R_i$, yielding an activation map. The new activation map plays as an information adaptor. It scores the part information contained in the trunk map $T_i$, by respecting the importance of parts. To avoid missing part information during the scoring process, we combine the residual map $D_i$ with the scored part information to form the context representation $F_i$. As shown in Fig.~\ref{fig:part_relation}, $F_i$ is fed to the next level for producing the context representation $F_{i+1}$.
Using Eq.~\ref{eq:context}, we compute context representations $\{F_i\}$ at all levels. Each level of context representation is used for scoring parts for the further selection process, as we will describe below.

\section{Part Selection Module}

The discriminative parts of hand bone are of importance to accurate estimation of bone age. Generally, recent methods use single or all parts for BAA, without understanding the part relation. As the relation model provide the importance of parts, the existing methods are incapable of selecting useful parts, which limits the performance of BAA. In this section, we advocate the idea of using the part relation to select useful parts.

\begin{figure*}[t!]
\centering
\includegraphics[width=0.75\linewidth]{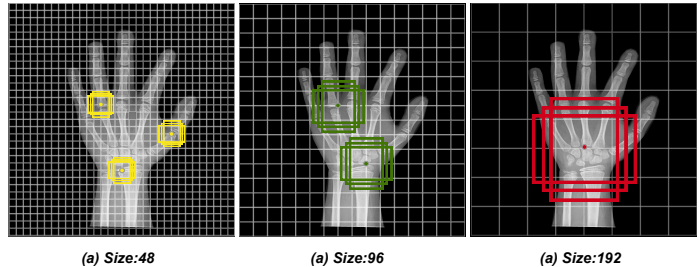}
\caption{Examples of visualized anchors in different scales and ratios.}
\vspace{-0.2in}
\label{fig:anchor}
\end{figure*}

We propose a part selection module to select the important parts, which are used to construct the representation for BAA. Given an X-ray image, our part selection module employs a set of anchors $\{A_n\}$ to represent the part candidates for selection. Note that anchors have different sizes and ratios, and span over the entire image to cover potential parts, as illustrated in Fig.~\ref{fig:anchor}.

Next, we conduct a scoring process on each anchor, as illustrated in Fig.~\ref{fig:overview}(b). This is done by using different levels of context representations, which are provided by the part relation module, to regress a set of scores $\{S_n\}$ for all anchors. Here, we resort to the convention of object detection~\cite{lin2017feature} and use lower/higher levels of representation to compute scores for smaller/larger anchors. Given the scores for all anchors, we sort the set of scores at all levels, then we employ the non-maximum suppression to eliminating overlapping anchors. and select top-M anchors having highest scores in the image. By following the top-M order of the selected anchors, we crop the corresponding image regions from the X-ray image. These image regions are resized uniformly, and each region is fed to another ResNet-34 for producing a feature vector. We sum all regions' feature vectors as a local feature. Together, we concatenate the local feature $F_{local}$, global feature $F_{global}$ (i.e., the highest level of context representation) and gender feature $F_{gender}$ (i.e., a one-hot vector) for predicting the bone age (see Fig.~\ref{fig:overview}(c)).

\begin{figure*}[t!]
	\centering
	\includegraphics[width=0.6\linewidth]{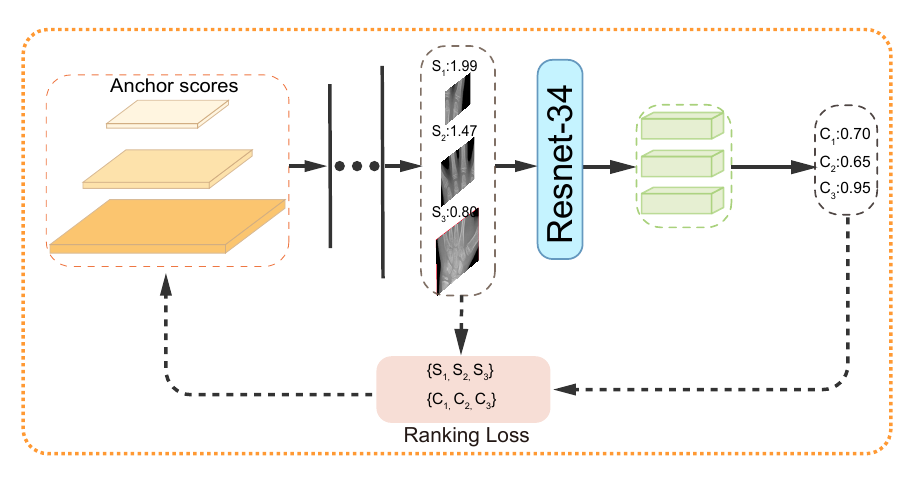}
	\caption{The ranking loss uses anchor scores and confidences to supervise the network training.}
    \vspace{-0.2in}
	\label{fig:rankingloss}
\end{figure*}

To jointly train the part relation and selection modules, we employ the ranking loss~\cite{yang2018learning} that measures the quality of part relation and selection. As illustrated in Fig.~\ref{fig:rankingloss}, given the selected parts, we associate them with anchor scores. We use each selected part to estimate a bone age, which is compared to the ground-truth age for computing a confidence. Here, the intuition is that the useful part, which has higher anchor score, contributes to the prediction of bone age that is similar to the ground-truth. Thus, the ranking loss for part relation and selection modules is computed as:
\begin{equation}
L_{rank}= \sum_{(i,j)}\mathbbm{1}(C_j>C_i)max((1-S_i-S_j),0),
\label{eq:eq4}
\end{equation}
where we denote $S_i$ as the anchor score for the $i^{th}$ selected part. Given the $i^{th}$ selected part, $C_i$ is the confidence formulated as:
\begin{equation}
C_i = 1-\sigma(-\left|y_i-y^*\right|),
\label{eq:eq2}
\end{equation}
where $\sigma$ also means the sigmoid activation function. $y_i$ means the predicted bone age by using the feature vector computed on $i^{th}$ selected part, and $y^*$ is the ground-truth bone age. The confidence measures the difference between the predicted age and the ground-truth age.
By using Eq.~\eqref{eq:eq4}, we compare each pair of parts. In the case where a part has higher anchor score but leads to lower confidence, by comparing to another part, the ranking loss proposes a penalty to guide the training of network.

%\di{Below needs rewriting...}
%However, the operation of cropping is differentiable, which causes the Part Selection Module unable to update parameters via standard gradient backpropagation.
%The ranking loss mainly consists of 2 parts: the information function and confidence function, the information function $I:A \rightarrow (-\infty, +\infty) $ is defined to evaluate how informative the parts are, and in here, we use the score of each anchor to measure theif information. and then confidence function $C:A \rightarrow [0, 1] $ is setted as a classifier to evaluate the confidence of part's representativeness, we use the Eq. \ref{eq2} to denote the confidence function.
%%\begin{equation}
%%C_i = 1-\sigma(-\left|y-y^*\right|)
%%\label{eq:eq2}
%%\end{equation}
%With the assumption that the more informative and useful part should have higher confidence, so the following condition should hold:
%\begin{equation} \label{eq3}
%\left\{(A_i, A_j),S_j > S_i \lvert C_j>C_i\right\}
%\end{equation}
%More informlly, we denote the Top-$M$ anchors as $\left\{A_1, A_2, ...A_m \right\}$, each achor will have corresponding score $S$, which represents it's informativeness, therefore, the ranking loss is defined as follow:
%
%the ranking loss stimulates  $I_i > I_j$ if the $C_i > C_j$, which provide direct guide for the regression of anchor score.
\section{Network Training}

%\di{What do you mean by joint or full logit? This below paragraph needs rewriting...}
%In this section, we will describe more details of the network training, as described in the Sec.3, the concatenated feature will regress the final predcition, and ranking loss will provide direct guidance to the anchor scores' updating, hence there are two not independent but rather complementary losses

For network training, we use an objective function, which includes the ranking loss for relation and section module and L2-loss for the final prediction of bone age. The objective function is formulated as:
\begin{equation} \label{eq4}
L_{total} = L_{rank} + \vert y - y^* \vert ^2.
\end{equation}

%We construct PRSNet with the open-source Pytorch toolkit~. We use different ResNet-34 models in the part relation and selection modules, respectively. In the part relation module, we employ the layers $res3$, $res4$ and $res5$ of the ResNet-34 model to compute anchor scores and select top-3 anchors over all levels. In the selection module, each anchor is used to extract a feature vector on the corresponding layer ($res3$, $res4$ or $res5$) of another ResNet-34 model. During the network training, we augment X-ray images with conventional strategies, i.e., horizontal flipping, rotation, shifting and scaling. We use SGD to optimize PRSNet, where each mini-batch consists of 16 $512 \times 512$ images. We set the initial learning rate to 1e-3, and decay the learning rate by 10 after every 25 epochs. We train the network for 100 epochs on 2 TITAN XP \textcolor{red}{in 5 hours totally. Compared to the baseline network, Our PRSNet 's inference time consumption is increased by 10\%, and the testing speed is 20 ms per image. }

We construct PRSNet with the open-source Pytorch toolkit~. We use different ResNet-34 models in the part relation and selection modules, respectively. In the part relation module, we employ the layers $res3$, $res4$ and $res5$ of the ResNet-34 model to compute anchor scores and select top-3 anchors over all levels. In the selection module, each anchor is used to extract a feature vector on the corresponding layer ($res3$, $res4$ or $res5$) of another ResNet-34 model. During the network training, we augment X-ray images with conventional strategies, i.e., horizontal flipping, rotation, shifting and scaling. We use SGD to optimize PRSNet, where each mini-batch consists of 16 $512 \times 512$ images. We set the initial learning rate to 1e-3, and decay the learning rate by 10 after every 25 epochs. We train the network on 2 TITAN XP, for 100 epochs that need 5 hours totally. The BAA process on a testing image requires 20 ms. On average, our PRSNet increases the training/testing time by 10\%, in comparison with the baseline network.

\section{Experiments}

\subsection{Dataset and Pre-processing}
We evaluate PRSNet on the dataset of RSNA 2017 Pediatric Bone Age Challenge. Totally, this dataset contains 12611 X-ray images for training and 200 images for testing. To reduce noise from background in X-ray images, we conduct a lightweight annotation on 200 images to train a foreground segmentation model. Below, we use the trained segmentation model to remove background and select regions of hand bones for training and testing PRSNet. We report all results on the test set, in terms of Mean Absolute Error (MAE). Smaller score of MAE means better performance.

\subsection{Ablation Studies}
%\vspace{-0.2in}
%\begin{table}
%	\small
%	\begin{center}
%	\begin{tabular}{l|c}
%		\hline
%		Method   & Mean Absolute Error \\ \hline
%		baseline &  6.52			   						\\ \hline
%		w/o part relation   &   5.05  	\\ \hline
%		w/o part selection  &   5.20  	\\ \hline	
%		PRSNet              &  \textbf{4.49}  	\\ \hline
%	\end{tabular}
%	\end{center}
%	\caption{Bone age assessment results on the RSNA Pediatric Bone Age Test Dataset.}
%	\vspace{-0.4in}
%	\label{tab:ablation}
%\end{table}

\vspace{-0.2in}
\begin{table}
	\begin{center}
	\begin{tabular}{l|c|c|c|c}
		\hline
		Method   & baseline & w/o relation & w/o selection & PRSNet \\ \hline
		MAE      &  6.52	&    5.05  	   &   5.20  	   &  \textbf{4.49}  	\\ \hline
	\end{tabular}
	\end{center}
	\caption{Bone age assessment results on the RSNA Pediatric Bone Age Test Dataset.}
	\vspace{-0.4in}
	\label{tab:ablation}
\end{table}

%\vspace{-0.15in}
First, we exam the effect on the BBA task by removing the key components of PRSNet, i.e., relation and selection modules. Without these modules, the whole model degrades to the baseline ResNet-34, which yields the score of 6.52 MAE. It lags far behind our full model in terms of BBA accuracy.
In Table~\ref{tab:ablation}, we test the network without using part relation maps. Here, the part relation module degrades to a basic ResNet-34 model, which yields the score of 5.05 MAE. Comparably, our full model achieves a better score of 4.49 MAE. This is because the relation module provides relevant part information, which is useful for constructing context representation.
Next, we disable the part selection module. In this case, we omit the local feature vector and only concatenate the global feature and gender feature for BAA. This model produces a score of 5.20 MAE, which is significantly lower than the result of our full model. Note that hand bone contains useless parts, which embed redundant information to the final feature vector. It demonstrates the effectiveness of our selection module in terms of choosing useful part information.

\begin{figure*}[t!]
	\centering
	\includegraphics[width=0.8\linewidth]{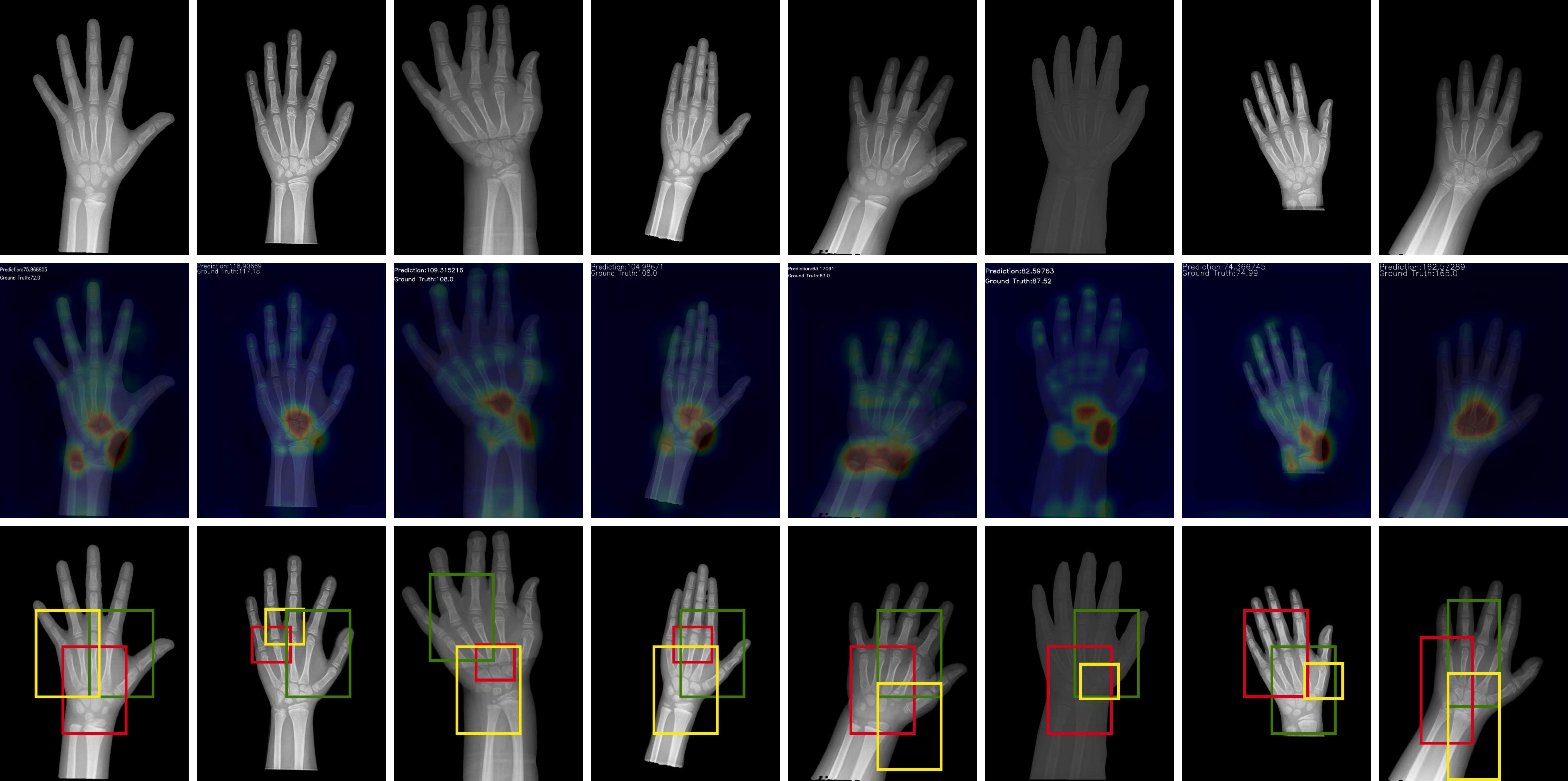}
	\caption{Visualization of part relation and selection results. Given the input images (fist row), the part relation module produce the relation maps (middle row) and the selection module produces the top-3 anchors (last row).}
    \vspace{-0.1in}
	\label{fig:results}
\end{figure*}

\subsection{Comparsion with State-of-the-Arts}

\begin{figure*}[t!]
	\centering
	\includegraphics[width=1.0\linewidth]{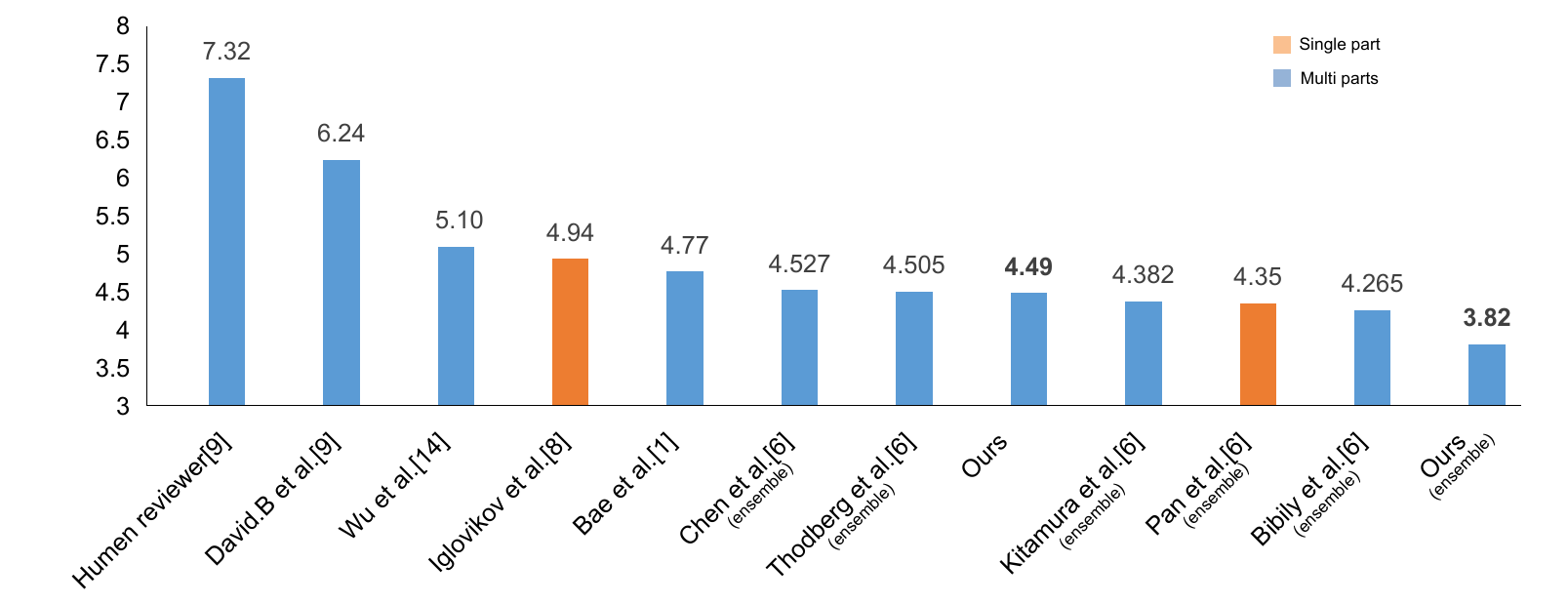}
	\caption{Comparison with different approaches on the testing set.}
    \vspace{-0.2in}
	\label{fig:compare}
\end{figure*}

In Fig~\ref{fig:compare}, we compare our PRSNet with state-of-the-art methods in terms of the accuracy of BAA. We divide the compared methods into two groups. In the first group, the methods (marked in yellow) select a single part for BAA. Compared to this kind of methods, our PRSNet yields better performance, since our approach provides richer context information of different parts.
In the second group, similar to our approach, the methods (marked in blue) also choose multiple parts for BAA. However, this kind of methods neither model part relation nor select parts according to their importance. Thus, our approach outperforms these methods. For a fair comparison with methods (see~\cite{halabi2018rsna} in Fig~\ref{fig:compare}) that ensemble several deep models, we train several PRSNets with different random initializations. We achieve a better result than other methods, demonstrating the effectiveness of our joint consideration of part relation and selection.

\section{Conclusion}
%In this paper, we presented the PRSNet include the Part Relation and Part Selection Module for automatic bone age assessment. The relatedness between the parts in X-images was enhanced and most useful parts are selected to further study. Extensive experiments on the benchmark dataset verified the efficacy of our method and achieved competitive performance over the state-of-the-arts.

Recent progress on bone age assessment benefits from context information of different parts. In this paper, we have proposed a novel scheme for modeling the part relation. Our method uses relation maps to activate concurrent parts, which form useful context information. Furthermore, the part relation improves the selection of parts for BAA, and the part selection module provides supervision for updating the relation model. We have demonstrated that our approach is effective and outperforms the state-of-the-art on the public benchmarks.

\section*{Acknowledgments}
We thank the anonymous reviewers for their constructive comments. This work was supported in part by NSFC (61702338) and Shenzhen Science and Technology Program (No.JCYJ20180507182410327).

\bibliographystyle{splncs04}
\bibliography{library}
\end{document}